\documentclass[twoside,11pt]{article}

\usepackage[preprint]{jmlr2e}
\usepackage[T1]{fontenc}

\usepackage{booktabs}
\usepackage{multirow}
\usepackage{makecell}
\usepackage[table]{xcolor}
\usepackage{pifont}
\usepackage{caption}
\usepackage{bm}

\definecolor{GithubBlue}{HTML}{1c88e3}
\definecolor{GithubGreen}{HTML}{7baf44}
\definecolor{GithubRed}{HTML}{e22d2d}
\definecolor{LightGrey}{HTML}{F0F0F0}

\newcommand{\cmark}{\textcolor{GithubGreen}{\ding{51}}}%
\newcommand{\xmark}{\textcolor{GithubRed}{\ding{55}}}%
\newcommand{\minus}{\textcolor{GithubBlue}{\rule[0.17em]{0.7em}{0.25em}}}

\newcommand{\imitation}{\textsc{imitation}}
\newcommand{\stablebaselines}{Stable Baselines3}

\newcommand{\pytorch}{PyTorch}

\newcommand{\testcoverage}{98\%}

\jmlrheading{1}{2022}{1-48}{4/00}{10/00}{meila00a}{Adam Gleave and Mohammad Taufeeque and Juan Rocamonde and Erik Jenner and Steven H. Wang and Nora Belrose and Sam Toyer and Scott Emmons and Stuart Russell}

\ShortHeadings{\imitation{}: Clean Imitation Learning Implementations}{Gleave et al.}
\firstpageno{1}

\begin{document}

    \title{\imitation{}: Clean Imitation Learning Implementations}
    \author{%
    \name Adam Gleave\textsuperscript{1,2}\thanks{Equal contribution}\ \thanks{Corresponding author} \email gleave@berkeley.edu \\
    \name Mohammad Taufeeque\textsuperscript{2}\footnotemark[1] \email 9taufeeque9@gmail.com \\
    \name Juan Rocamonde\textsuperscript{2}\footnotemark[1] \email juan@alignmentfund.org \\
    \name Erik Jenner\textsuperscript{1} \email jenner@berkeley.edu \\
    \name Steven H. Wang\textsuperscript{3} \email sh.wang@berkeley.edu \\
    \name Sam Toyer\textsuperscript{1} \email sdt@berkeley.edu \\
    \name Maximilian Ernestus\textsuperscript{1,2} \email maximilian@ernestus.com \\
    \name Nora Belrose\textsuperscript{2} \email nora@alignmentfund.org \\
    \name Scott Emmons\textsuperscript{1,2} \email emmons@berkeley.edu \\
    \name Stuart Russell\textsuperscript{1} \email russell@cs.berkeley.edu
    \AND
    \addr 
    \textsuperscript{1}UC Berkeley
    \textsuperscript{2}FAR AI
    \textsuperscript{3}ETH Zurich
    }

    \editor{Editor Name}

    \maketitle

    \begin{abstract}
        \imitation{} provides open-source implementations of imitation and reward learning algorithms in \pytorch{}.
        We include three inverse reinforcement learning (IRL) algorithms, three imitation learning algorithms and a preference comparison algorithm.
        The implementations have been benchmarked against previous results, and automated tests cover \testcoverage{} of the code.
        Moreover, the algorithms are implemented in a modular fashion, making it simple to develop novel algorithms in the framework.
        Our source code, including documentation and examples, is available at \url{https://github.com/HumanCompatibleAI/imitation}.
    \end{abstract}

    \begin{keywords}
        Imitation Learning,
        Reward Learning,
        Software,
        Python,
        PyTorch
    \end{keywords}

    \section{Introduction}
    Reinforcement learning (RL) has surpassed human performance in domains with clearly-defined reward functions, such as games~\citep{berner2019dota}.
    Unfortunately, it is difficult or impossible to procedurally specify the reward function for many real-world tasks.
    We must instead \emph{learn} a reward function or policy directly from user feedback.
    Moreover, even when we can write down a reward function, such as if the agent wins a game, the resulting objective might be so sparse that RL cannot efficiently solve it.
    State-of-the-art results in RL therefore often use imitation learning to initialize the policy~\citep{oriol2019alphastar}.
    
    We introduce \imitation{}: a library providing high-quality, reliable and modular implementations of seven reward and imitation learning algorithms.
    Crucially, our algorithms follow a consistent interface, making it simple to train and compare a range of algorithms.
    Furthermore, \imitation{} is built using modern backends such as \pytorch{} and \stablebaselines{}.
    By contrast, prior libraries typically support only a handful of algorithms, are no longer actively maintained, and are built on top of deprecated frameworks.
    
    A key use case of \imitation{} is as an experimental \emph{baseline}.
    Prior work has shown that small implementation details in imitation learning algorithms can have significant impacts on performance~\citep{orsini2021matters}.
    This could lead to spurious positive results being reported if a weak experimental baseline were used.
    To address this challenge, our algorithms have been carefully benchmarked and compared to prior implementations (see Figure~\ref{fig:results} and Table~\ref{tab:results}).
    Additionally, our test suite covers \testcoverage{} of our code, and we also perform static type checking.
    
    In addition to providing reliable baselines, \imitation{} aims to simplify developing novel reward and imitation learning algorithms.
    Our implementations are \emph{modular}: users can freely change the reward or policy network architecture, RL algorithm and optimizer without any changes to the code.
    Algorithms can be extended by subclassing and overriding the relevant methods.
    Moreover, to support the development of entirely novel algorithms, \imitation{} provides utility methods to handle common tasks such as collecting rollouts.
    
    
    
    \section{Features}
    \label{sec:features}
    
    \paragraph{Comprehensive}
    \imitation{} implements seven algorithms spanning a range of reward and imitation learning styles.
    Our IRL algorithms consist of 1) the seminal tabular method Maximum Causal Entropy IRL~\citep[MCE IRL;][]{ziebart2010mceirl}, 2) a baseline based on density estimation, and 3) the state-of-the-art approach Adversarial IRL~\citep[AIRL;][]{fu2018airl}.
    For imitation learning, we include 1) the simple Behavioral Cloning (BC) algorithm, 2) a variant DAgger~\citep{ross2011} that learns from interactive demonstrations, and 3) the state-of-the-art Generative Advesarial Imitation Learning~\citep{ho2016gail} algorithm.
    Finally, we also include Deep RL from Human Preferences~\citep[DRLHP;][]{christiano2017drlhp} that infers a reward function from comparisons between trajectory fragments.
    
    \paragraph{Consistent Interface}
    We provide a unified API for all algorithms, inheriting from a common base class \texttt{BaseImitationAlgorithm}.
    Algorithms diverge only where strictly necessary (e.g. a different feedback modality).
    This makes it simple to automatically test a wide range of algorithms against a benchmark suite.
    
    \paragraph{Experimental Framework}
    We provide scripts to train and evaluate the algorithms, making it easy to use the library without writing a single line of code.
    The scripts follow a consistent interface, and we include examples to run all algorithms on a suite of commonly used environments.
    To ensure replicable experiments we use Sacred~\citep{greff2017sacred} for configuration and logging.
    
    \paragraph{Modularity}
    To support the variety of use cases that arise in research, we have designed our implementations to be modular and highly configurable.
    For example, algorithms can be configured to use any of the seven \stablebaselines{} RL algorithms (or a custom algorithm matching this interface).
    By contrast, prior implementations often implemented imitation learning algorithms by subclassing a specific RL algorithm, requiring substantial code modification to be ported to new RL algorithms.

    We have also designed the code to be easy to extend in order to implement novel algorithms.
    Each algorithm is implemented by a class with instance methods corresponding to each logical step of the algorithm.
    New algorithms can be implemented simply by subclassing an existing algorithm and overriding a subset of methods.
    This power is illustrated by our implementations of GAIL and AIRL, which both subclass \texttt{AdversarialTrainer}.
    They differ only in the choice of discriminator, with most training logic shared.
    
    \paragraph{Documentation}
    \imitation{} comes with extensive documentation available at \url{https://imitation.readthedocs.io}.
    We include installation instructions, a quickstart guide and a contribution guide for prospective developers as well as an API reference.
    We also provide tips for evaluation of imitation and reward learning algorithms, including avoiding variable-horizon environments which has confounded prior evaluation~\citep{kostrikov2018bias}.
    
    \begin{figure}[t]
        \begin{center}
        \includegraphics{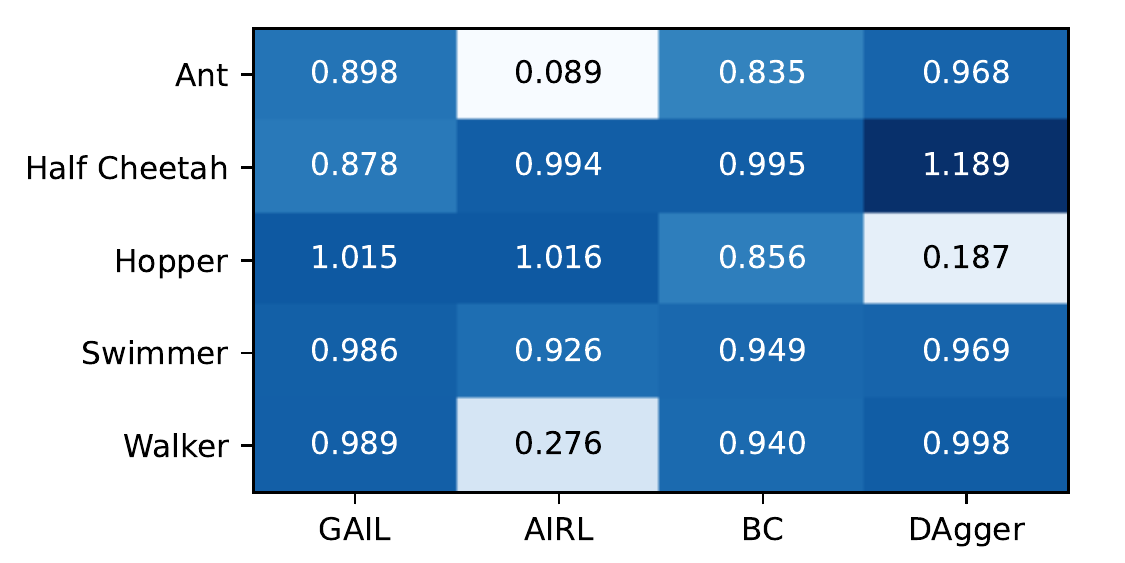}
        \end{center}
        \caption{Returns of our algorithms normalized so that $1$ is the returns of an expert policy and $0$ is that of a random policy. Our algorithms reach close to expert performance on most environments. Detailed results, including confidence intervals, can be found in Table~\ref{tab:results}.}
        \label{fig:results}
    \end{figure}

    \paragraph{High-Quality Implementations}
    We take great care to provide reliable implementations of algorithms.
    Our test suite covers \testcoverage{} of the entire codebase. Additionally, we use type annotations throughout, and statically check our code using \texttt{pytype} and \texttt{mypy}.
    
    While our thorough testing and code review help avoid bugs, even apparently minor implementation details can have significant impacts on algorithm performance~\citep{engstrom2020implementation}.
    Therefore, we have also benchmarked our algorithms on environments that have been commonly used in prior work, including in the original papers of the algorithms.

We find in Figure~\ref{fig:results} that our algorithms reach expert-level performance on these environments, with the exception of AIRL in the Ant and Walker environments, and DAgger in the Hopper environment. AIRL and DAgger were not originally tested on the Walker and Hopper environments, respectively, so it is possible these algorithms just do not perform well on these environments. The AIRL paper did report positive results on an Ant environment, whereas our implementation performs close to random. However, the AIRL paper used a custom version of the Ant environment, whereas we use the standard Gym environment (see Table~\ref{tab:envs} for a description of the environments used for benchmarking).







    \section{Comparison to Other Software}

    A key advantage of \imitation{} is the breadth of reward and imitation learning algorithms implemented. 
    \imitation{} includes a total of seven algorithms, whereas Table~\ref{tab:library-comparison} shows most other software packages include only one or two.
    This broad coverage allows users to easily test a large number of baselines, without needing to find and integrate multiple libraries.
    
    Another benefit of \imitation{} is that it is built on modern frameworks like \pytorch{} and \stablebaselines{}.
    By contrast, many extant implementations of imitation and reward learning algorithms were released many years ago and have not been actively maintained.
    This is particularly true for reference implementations released with original papers, such as the GAIL~\citep{ho2016gailcode} and AIRL~\citep{fu2018airlcode} codebases.
    However, even popular libraries like Stable Baselines2 are no longer under active development\footnote{The successor to Stable Baselines2, \stablebaselines{}, has dropped support for imitation algorithms in favour of \imitation{}'s own implementation~\citep{sb2020imitation}.}.
    
    We compare alternative libraries on a variety of metrics in Table~\ref{tab:library-comparison}.
    Although it is not feasible to include every implementation of imitation and reward learning algorithms, to the best of our knowledge this table includes all widely-used imitation learning libraries.
    We find that \imitation{} equals or surpasses alternatives in all metrics.
    APRel~\citep{biyik2021aprel} also scores highly but focuses on preference comparison algorithms learning from low-dimensional features.
    This is complementary to \imitation{}, which provides a broader range of algorithms and emphasizes scalability, at the cost of greater implementation complexity.


    \begin{table}[!ht]
        \renewcommand{\arraystretch}{1.2}
        \newcommand{\coveragesymbol}{**}
        \newcommand{\narlsymbol}{\textdagger}
        \newcommand{\trpoonlysymbol}{\S}
        \newcommand{\configoptimizersymbol}{\P}
        \newcommand{\aprelsymbol}{*}
        \centering
        \rowcolors{2}{white}{LightGrey}
        \newcommand{\rotheading}[1]{\rotatebox{0}{#1}}
        \setlength{\tabcolsep}{4pt}
        \begin{tabular}{@{}lccccccc@{}}
            \toprule
             & \rotheading{imitation} & \rotheading{APReL} & \rotheading{\makecell{OpenAI \\ Baselines}} & \rotheading{\makecell{Stable \\ Baselines2}} & \rotheading{\makecell{Intel \\ COACH}} & \rotheading{\makecell{GAIL \\ Paper}} & \rotheading{\makecell{AIRL \\ Paper}} \\
            \midrule
            Backend & PyTorch & NumPy & TF1 & TF1 & TF1/MxNet & Theano & TF1 \\
            \makecell[l]{\# imitation \\ algorithms} & 7 & 1\textsuperscript{\aprelsymbol} & 1 & 1 & 2 & 2 & 4 \\ 
            \makecell[l]{Last Commit \hspace*{1.95em}\\ (age)} & \textless 1w & \textless 1m & >2.5y & \textgreater 3m & 1m
                & \textgreater 4y & \textgreater 4y \\
            \makecell[l]{Approved PRs \\ (6 months)} & 103 & 0 & 0 & 1 & 7 & 0 & 0 \\
            PEP8 & \cmark & \xmark & \cmark & \cmark & \xmark & \xmark & \xmark \\
            Type Annotations & \cmark & \cmark & \xmark & \cmark & \cmark & \xmark & \xmark \\
            Type Checking & \cmark & \xmark & \xmark & \cmark & \xmark & \xmark & \xmark \\
            Test Coverage & \testcoverage{} & \xmark & 49\%\textsuperscript{\coveragesymbol} & 89\%\textsuperscript{\coveragesymbol} & \textgreater 58\%\textsuperscript{\coveragesymbol} &
            \xmark & \xmark \\
            Documentation & \cmark & \cmark & \xmark & \cmark & \cmark & \xmark & \xmark \\
            Custom RL Agent & \cmark & N/A\textsuperscript{\narlsymbol} & \xmark\textsuperscript{\trpoonlysymbol} & \xmark\textsuperscript{\trpoonlysymbol} & N/A\textsuperscript{\narlsymbol} & \xmark\textsuperscript{\trpoonlysymbol} & \xmark\textsuperscript{\trpoonlysymbol} \\
            Custom Optimizer & \cmark & \cmark & \xmark & \xmark & \minus\textsuperscript{\configoptimizersymbol} & \xmark & \cmark \\
            \bottomrule
        \end{tabular}
        \caption{\imitation{} compares favourably to alternative libraries in terms of number of imitation learning algorithms implemented, project activity, implementation quality and flexibility. Only \imitation{}, APReL and COACH use modern backends. \\
        \\
        \textbf{Key}: \textbf{\aprelsymbol} Is a single Bayesian algorithm, but supports different feedback formats (e.g. preference ranking and comparisons) and methods for querying feedback. \textbf{\coveragesymbol} coverage not officially reported, estimated by us from running test suite; \textbf{\narlsymbol} does not use RL; \textbf{\trpoonlysymbol} TRPO is the only RL algorithm supported; \textbf{\configoptimizersymbol} configurable but limited to Adam, RMSProp and LBFGS.}


        \label{tab:library-comparison}
    \end{table}


    \acks{Thanks to open-source contributors who have reported bugs, feature enhancements or made code contributions. We would in particular like to thank Yawen Duan, Lev McKinney, Nevan Wichers, Dan Pandori, Tom Tseng, Yulong Lin, Ian Fan, Ansh Radhakrishnan and Samuel Arnesen for their code contributions.}

    \section*{Author contributions}
    Adam Gleave managed the project, performed code reviews and made a variety of minor code contributions. Mohammad Taufeeque benchmarked the algorithms, improved the documentation and made other minor code contributions. Juan Rocamonde edited the manuscript, added MyPy typing support, and made other minor code improvements. Erik Jenner added the initial implementation of our preference comparison algorithm. Steven H. Wang was the primary developer of the original, TensorFlow codebase. Nora Belrose improved the documentation, added new algorithmic features, and made minor code improvements. Sam Toyer implemented initial versions of several algorithms and assisted with the PyTorch port. Scott Emmons led the initial port to PyTorch and \stablebaselines{}. Stuart Russell provided research advice.

    \bibliography{refs}

\begin{thebibliography}{16}
\providecommand{\natexlab}[1]{#1}
\providecommand{\url}[1]{\texttt{#1}}
\expandafter\ifx\csname urlstyle\endcsname\relax
  \providecommand{\doi}[1]{doi: #1}\else
  \providecommand{\doi}{doi: \begingroup \urlstyle{rm}\Url}\fi

\bibitem[Berner et~al.(2019)Berner, Brockman, Chan, Cheung, Dębiak, Dennison,
  Farhi, Fischer, Hashme, Hesse, Józefowicz, Gray, Olsson, Pachocki, Petrov,
  d.~O.~Pinto, Raiman, Salimans, Schlatter, Schneider, Sidor, Sutskever, Tang,
  Wolski, and Zhang]{berner2019dota}
Christopher Berner, Greg Brockman, Brooke Chan, Vicki Cheung, Przemysław
  Dębiak, Christy Dennison, David Farhi, Quirin Fischer, Shariq Hashme, Chris
  Hesse, Rafal Józefowicz, Scott Gray, Catherine Olsson, Jakub Pachocki,
  Michael Petrov, Henrique~P. d.~O.~Pinto, Jonathan Raiman, Tim Salimans,
  Jeremy Schlatter, Jonas Schneider, Szymon Sidor, Ilya Sutskever, Jie Tang,
  Filip Wolski, and Susan Zhang.
\newblock Dota 2 with large scale deep reinforcement learning, 2019.
\newblock arXiv:1912.06680 [cs.LG].

\bibitem[Bıyık et~al.(2021)Bıyık, Talati, and Sadigh]{biyik2021aprel}
Erdem Bıyık, Aditi Talati, and Dorsa Sadigh.
\newblock {APReL}: A library for active preference-based reward learning
  algorithms, 2021.

\bibitem[Christiano et~al.(2017)Christiano, Leike, Brown, Martic, Legg, and
  Amodei]{christiano2017drlhp}
Paul~F Christiano, Jan Leike, Tom Brown, Miljan Martic, Shane Legg, and Dario
  Amodei.
\newblock Deep reinforcement learning from human preferences.
\newblock In \emph{Conference on Neural Information Processing Systems
  (NeurIPS)}, pages 4299--4307, 2017.

\bibitem[Engstrom et~al.(2020)Engstrom, Ilyas, Santurkar, Tsipras, Janoos,
  Rudolph, and Madry]{engstrom2020implementation}
Logan Engstrom, Andrew Ilyas, Shibani Santurkar, Dimitris Tsipras, Firdaus
  Janoos, Larry Rudolph, and Aleksander Madry.
\newblock Implementation matters in deep {RL}: A case study on {PPO} and
  {TRPO}.
\newblock In \emph{International Conference on Learning Representations
  (ICLR)}, 2020.

\bibitem[Fu(2018)]{fu2018airlcode}
Justin Fu.
\newblock Inverse {RL}.
\newblock \url{https://github.com/justinjfu/inverse_rl}, 2018.

\bibitem[Fu et~al.(2018)Fu, Luo, and Levine]{fu2018airl}
Justin Fu, Katie Luo, and Sergey Levine.
\newblock Learning robust rewards with adverserial inverse reinforcement
  learning.
\newblock In \emph{International Conference on Learning Representations}, 2018.

\bibitem[Gleave et~al.(2020)Gleave, Freire, Wang, and Toyer]{gleave2020seals}
Adam Gleave, Pedro Freire, Steven Wang, and Sam Toyer.
\newblock {seals}: Suite of environments for algorithms that learn
  specifications.
\newblock \url{https://github.com/HumanCompatibleAI/seals}, 2020.

\bibitem[Greff et~al.(2017)Greff, Klein, Chovanec, Hutter, and
  Schmidhuber]{greff2017sacred}
Klaus Greff, Aaron Klein, Martin Chovanec, Frank Hutter, and J{\"u}rgen
  Schmidhuber.
\newblock The {Sacred} infrastructure for computational research.
\newblock In \emph{Python in Science}, volume~28, pages 49--56, 2017.

\bibitem[Ho and Ermon(2016)]{ho2016gail}
Jonathan Ho and Stefano Ermon.
\newblock Generative adversarial imitation learning.
\newblock In \emph{Advances in Neural Information Processing Systems}, pages
  4565--4573, 2016.

\bibitem[Ho and Hesse(2016)]{ho2016gailcode}
Jonathan Ho and Christopher Hesse.
\newblock Generative adversarial imitation learning.
\newblock \url{https://github.com/openai/imitation}, 2016.

\bibitem[Kostrikov et~al.(2019)Kostrikov, Agrawal, Dwibedi, Levine, and
  Tompson]{kostrikov2018bias}
Ilya Kostrikov, Kumar~Krishna Agrawal, Debidatta Dwibedi, Sergey Levine, and
  Jonathan Tompson.
\newblock Discriminator-actor-critic: Addressing sample inefficiency and reward
  bias in adversarial imitation learning.
\newblock In \emph{International Conference on Learning Representations
  (ICLR)}, 2019.

\bibitem[Orsini et~al.(2021)Orsini, Raichuk, Hussenot, Vincent, Dadashi,
  Girgin, Geist, Bachem, Pietquin, and Andrychowicz]{orsini2021matters}
Manu Orsini, Anton Raichuk, Léonard Hussenot, Damien Vincent, Robert Dadashi,
  Sertan Girgin, Matthieu Geist, Olivier Bachem, Olivier Pietquin, and Marcin
  Andrychowicz.
\newblock What matters for adversarial imitation learning?, 2021.
\newblock arXiv:2106.00672 [cs.LG].

\bibitem[Raffin et~al.(2021)Raffin, Hill, Ernestus, Gleave, Kanervisto, and
  Dormann]{sb2020imitation}
Antonin Raffin, Ashley Hill, Maximilian Ernestus, Adam Gleave, Anssi
  Kanervisto, and Noah Dormann.
\newblock Imitation learning.
\newblock
  \url{https://stable-baselines3.readthedocs.io/en/master/guide/imitation.html},
  2021.

\bibitem[Ross et~al.(2011)Ross, Gordon, and Bagnell]{ross2011}
St{\'e}phane Ross, Geoffrey Gordon, and Drew Bagnell.
\newblock A reduction of imitation learning and structured prediction to
  no-regret online learning.
\newblock In \emph{International Conference on Artificial Intelligence and
  Statistics (AISTATS)}, pages 627--635, 2011.

\bibitem[Vinyals et~al.(2019)Vinyals, Babuschkin, Czarnecki, Mathieu, Dudzik,
  Chung, Choi, Powell, Ewalds, Georgiev, et~al.]{oriol2019alphastar}
Oriol Vinyals, Igor Babuschkin, Wojciech~M Czarnecki, Micha{\"e}l Mathieu,
  Andrew Dudzik, Junyoung Chung, David~H Choi, Richard Powell, Timo Ewalds,
  Petko Georgiev, et~al.
\newblock Grandmaster level in {StarCraft II} using multi-agent reinforcement
  learning.
\newblock \emph{Nature}, 575\penalty0 (7782):\penalty0 350--354, 2019.

\bibitem[Ziebart et~al.(2010)Ziebart, Bagnell, and Dey]{ziebart2010mceirl}
Brian~D. Ziebart, J.~Andrew Bagnell, and Anind~K. Dey.
\newblock Modeling interaction via the principle of maximum causal entropy.
\newblock In \emph{International Conference on Machine Learning (ICML)}, pages
  1255--1262, 2010.

\end{thebibliography}
    \clearpage
    \appendix
    \section{Detailed benchmarking results}
    The table below contains a more detailed account of the results obtained during benchmarking, including confidence intervals. The hyperparameters used to train these algorithms can be found on the \texttt{benchmarking} folder of the v0.3.2 release of imitation.
    \begin{table}[!ht]
        \begin{center}
    \begin{tabular}{llllll}
\toprule
{} &             Ant &    Half Cheetah &         Hopper &      Swimmer &          Walker \\
\midrule
Random &   $-356 \pm 22$ &   $-272 \pm 36$ &   $-60 \pm 59$ &    $1 \pm 7$ &    $-22 \pm 63$ \\
Expert &  $2408 \pm 110$ &  $3465 \pm 162$ &  $2631 \pm 19$ &  $298 \pm 1$ &  $2673 \pm 112$ \\
GAIL   &  $2126 \pm 119$ &  $3010 \pm 157$ &  $2671 \pm 27$ &  $294 \pm 2$ &   $2643 \pm 98$ \\
AIRL   &   $-110 \pm 37$ &  $3444 \pm 189$ &  $2675 \pm 24$ &  $276 \pm 6$ &    $722 \pm 81$ \\
BC     &  $1953 \pm 123$ &  $3446 \pm 130$ &  $2243 \pm 20$ &  $283 \pm 1$ &   $2512 \pm 86$ \\
DAgger &  $2321 \pm 127$ &  $4172 \pm 104$ &    $442 \pm 9$ &  $289 \pm 2$ &  $2669 \pm 110$ \\
\bottomrule
\end{tabular}
        \end{center}
    \caption{Benchmarking results for different algorithms and environments, in addition to random and expert policies. Each cell contains the estimate of the mean returns and a 95\% confidence interval for a t-distribution $t_{n-1}(\hat\mu,\,\hat\sigma^2,\,n-1)$, where $\hat \mu$ is the mean returns, $\hat \sigma^2$ is an estimate of the variance of $\hat \mu$, and $n$ is the number of experiments that were run, each with different seeds. In our benchmarks, $n=5$, and for each experiment,  $\approx 50$ trajectories were collected.}
    \label{tab:results}
    \end{table}

    \section{Environments used for benchmarking}
    All the environments used for benchmarking are standard Gym environments with some modifications to the default configuration. These ship within the Seals package \citep{gleave2020seals}. MuJoCo environments from Seals are configured to always include position in the observation and prevent early termination when environments are unhealthy. The Half Cheetah and Swimmer environments naturally do not terminate early, whereas the Hopper, Ant, and Walker environments are explicitly configured to behave in this way. 
    
    The table below lists the Gym IDs of the environments, and can be loaded by first installing Seals from PyPI.

    \begin{table}[h]
        \centering
        \begin{tabular}{lll}
        \toprule
             Name & ID & Base \\
             \midrule
             Ant & \texttt{seals/Ant-v0} & \texttt{Ant-v3} \\
             Half Cheetah & \texttt{seals/HalfCheetah-v0} & \texttt{HalfCheetah-v3} \\
             Hopper & \texttt{seals/Hopper-v0} & \texttt{Hopper-v3} \\
             Swimmer & \texttt{seals/Swimmer-v0} & \texttt{Swimmer-v3} \\
             Walker & \texttt{seals/Walker2d-v0} & \texttt{Walker2d-v3} \\
             \bottomrule
        \end{tabular}
        \caption{Environments used for benchmarking, including the Gym ID, and the Gym ID of the base environment (from \texttt{gym.envs.mujoco}).}
        \label{tab:envs}
    \end{table}

\end{document}